\documentclass[conference]{IEEEtran}
\IEEEoverridecommandlockouts
\usepackage{cite}
\usepackage{amsmath,amssymb,amsfonts}
\usepackage{algorithmic}
\usepackage{graphicx}
\usepackage{textcomp}
\usepackage{xcolor}
\usepackage{float}
\usepackage{tikz}
\usetikzlibrary{shapes.geometric, arrows}

\def\BibTeX{{\rm B\kern-.05em{\sc i\kern-.025em b}\kern-.08em
    T\kern-.1667em\lower.7ex\hbox{E}\kern-.125emX}}
\begin{document}

\title{Hybrid CNN with Chebyshev Polynomial Expansion for Medical Image Analysis}

\author{\IEEEauthorblockN{1\textsuperscript{st} Abhinav Roy}
\IEEEauthorblockA{\textit{dept. of DSAI} \\
\textit{IIIT, Naya Raipur}\\
Raipur, India \\
abhinav21102@iiitnr.edu.in}
\and
\IEEEauthorblockN{2\textsuperscript{nd} Bhavesh Gyanchandani}
\IEEEauthorblockA{\textit{dept. of DSAI} \\
\textit{IIIT, Naya Raipur}\\
Raipur, India \\
gyanchandani21102@iiitnr.edu.in}
\and
\IEEEauthorblockN{3\textsuperscript{rd} Aditya Oza}
\IEEEauthorblockA{\textit{dept. of DSAI} \\
\textit{IIIT, Naya Raipur}\\
Raipur, India \\
aditya21102@iiitnr.edu.in}

}

\maketitle

\begin{abstract}
Lung cancer remains one of the leading causes of cancer-related mortality worldwide, with early and accurate diagnosis playing a pivotal role in improving patient outcomes. Automated detection of pulmonary nodules in computed tomography (CT) scans is a challenging task due to variability in nodule size, shape, texture, and location. Traditional Convolutional Neural Networks (CNNs) have shown considerable promise in medical image analysis; however, their limited ability to capture fine-grained spatial-spectral variations restricts their performance in complex diagnostic scenarios. In this study, we propose a novel hybrid deep learning architecture that incorporates Chebyshev polynomial expansions into CNN layers to enhance expressive power and improve the representation of underlying anatomical structures. The proposed Chebyshev-CNN leverages the orthogonality and recursive properties of Chebyshev polynomials to extract high-frequency features and approximate complex nonlinear functions with greater fidelity. The model is trained and evaluated on benchmark lung cancer imaging datasets, including LUNA16 and LIDC-IDRI, achieving superior performance in classifying pulmonary nodules as benign or malignant. Quantitative results demonstrate significant improvements in accuracy, sensitivity, and specificity compared to traditional CNN-based approaches. This integration of polynomial-based spectral approximation within deep learning provides a robust framework for enhancing automated medical diagnostics and holds potential for broader applications in clinical decision support systems.
\end{abstract}


\section{Introduction}

Medical imaging plays a crucial role in the early detection and diagnosis of various diseases, enabling clinicians to make timely and accurate decisions. Deep learning, particularly Convolutional Neural Networks (CNNs), has significantly advanced the field by automating image analysis and improving classification accuracy \cite{suzuki2017overview}. Despite these advancements, challenges such as data scarcity, overfitting, and interpretability remain key issues in medical image analysis. To address these challenges, novel architectural modifications and mathematical enhancements in deep learning models have been explored \cite{oza2024hybrid, roy2024convkan}.

One promising direction in deep learning is the integration of mathematical functions to enhance feature extraction and model efficiency. Chebyshev polynomials, known for their strong approximation properties \cite{mason2002chebyshev}, have been successfully applied in various image processing tasks, including image fusion \cite{omar2010two}. However, their potential within deep learning architectures for medical imaging remains underexplored. Inspired by recent advancements in mathematical modeling and deep learning, we propose a novel CNN architecture that incorporates Chebyshev polynomials to improve feature extraction and classification performance in medical image analysis.

Recent innovations such as ConvKAN and TriSpectraKAN have demonstrated the benefits of embedding mathematical formulations like Kolmogorov-Arnold Networks and spectral decomposition into neural network frameworks to enhance interpretability and accuracy in tasks like brain tumor and COPD detection \cite{roy2024convkan, roy2025trispectrakan}. These approaches highlight a growing trend in using theoretically grounded functions within deep learning pipelines to address domain-specific challenges in medical imaging.

Lung cancer remains one of the leading causes of mortality worldwide, emphasizing the need for accurate and efficient diagnostic tools. Traditional imaging modalities, such as computed tomography (CT) and magnetic resonance imaging (MRI), have been extensively used for lung cancer detection. However, manual analysis is time-consuming and prone to errors. Deep learning-based approaches have demonstrated promising results in automating lung cancer detection \cite{wang2022deep}, yet challenges such as class imbalance and feature representation persist. By leveraging Chebyshev polynomial expansions within CNNs, our proposed method aims to enhance feature extraction capabilities, leading to improved classification accuracy while maintaining computational efficiency.


This paper presents a novel Chebyshev polynomial-based CNN for lung cancer classification. Our key contributions are as follows:  
\begin{itemize}
    \item We introduce a CNN architecture that integrates Chebyshev polynomials for enhanced feature extraction.  
    \item We evaluate our model on lung cancer datasets and compare its performance against traditional CNN-based approaches.  
    \item We demonstrate the effectiveness of mathematical function integration in deep learning models for medical imaging.  
\end{itemize}  

The rest of this paper is structured as follows: Section II discusses related works, highlighting previous advancements in medical imaging and mathematical function-based deep learning models. Section III details our proposed methodology, including network architecture and data preprocessing. Section IV presents experimental results and analysis. Finally, Section V concludes the paper with insights into future research directions.  
\section{Related Works}

Deep learning techniques have significantly advanced the field of medical imaging, particularly in lung cancer detection. Several studies have explored convolutional neural networks (CNNs), spectral learning, and mathematical function-based enhancements to improve classification accuracy and reduce false positives in high-stakes diagnostic tasks.

\subsection{Deep Learning in Medical Imaging}

Deep learning models, particularly CNNs, have demonstrated remarkable success in medical image analysis. Suzuki \cite{suzuki2017overview} provides a comprehensive overview of deep learning applications in medical imaging, highlighting the progression of CNN-based architectures in clinical domains. Kayalibay et al. \cite{kayalibay2017cnn} emphasized the advantage of 3D CNNs over traditional 2D models for segmentation tasks involving volumetric medical data. Similarly, Wang \cite{wang2022deep} showcased the use of AI-driven systems in automating lung cancer detection and reducing diagnostic delays.

Numerous efforts have focused on refining CNN architectures to improve performance. Dou et al. \cite{dou2016multilevel} proposed a multi-level 3D CNN that leveraged contextual information to reduce false positives in pulmonary nodule detection. Setio et al. \cite{setio2016pulmonary} introduced a 2D multi-view CNN that considered nodule characteristics from multiple angles, enhancing sensitivity. Eun et al. \cite{eun2018single} designed a compact 2D single-view model that attained a strong balance between accuracy and efficiency in CT-based detection.

\subsection{Mathematical Approaches in Image Processing}

Mathematical transforms and polynomials have long been used for image processing and analysis. Chebyshev polynomials, in particular, have proven effective for feature approximation and image fusion. Mason and Handscomb \cite{mason2002chebyshev} detail their mathematical foundations and use in numerical approximations. Omar et al. \cite{omar2010two} applied 2D Chebyshev polynomials for image fusion, outperforming conventional methods like ICA and dual-tree complex wavelet transforms (DT-CWT), especially in noisy environments.

Modern works such as ProKAN \cite{gyanchandani2024prokan} and TriSpectraKAN \cite{roy2025trispectrakan} reveal the benefits of integrating polynomial approximators and spectral learning into neural networks for segmentation and respiratory disease detection, respectively. These works support the efficacy of mathematically inspired architectures in biomedical image analysis.

\subsection{Lung Cancer Detection using AI}

Several deep learning-based models have been proposed for early lung cancer diagnosis. Jin et al. \cite{jin2017learning} introduced a 3D CNN for detecting malignancies in low-dose CT (LDCT) scans, achieving competitive accuracy on the KDSB17 dataset. Zhu et al. \cite{zhu2018deeplung} presented DeepLung, a 3D dual-path CNN capable of both detection and classification of pulmonary nodules, with strong results on LUNA16 and LIDC-IDRI datasets. Similarly, Zhang et al. \cite{zhang2018nodule} proposed NODULe, an advanced detection framework achieving a high CPM score and sensitivity.

Real-time and lightweight models have also gained traction. George et al. \cite{george2018using} adapted the YOLO framework for fast lung nodule detection, while Serj et al. \cite{serj2018deep} and Schwyzer et al. \cite{schwyzer2018automated} explored deep CNNs and DNNs for diagnostics using CT and PET modalities.

\subsection{Hybrid Approaches and Future Directions}

Hybrid models that combine CNNs with other signal processing or mathematical techniques have gained popularity. Theresa and Bharathi \cite{theresa2016cad} employed shearlet and complex wavelet transforms in CAD systems for lung nodule detection, reporting 96\% accuracy. More recently, Oza et al. \cite{oza2024hybrid} introduced a hybrid CNN-LSTM framework using QRS detection for heart failure classification. Roy et al. \cite{roy2024convkan} proposed ConvKAN, combining CNNs with Kolmogorov-Arnold Networks (KANs), improving classification in brain tumor MRI data.

These findings suggest that blending neural networks with principled mathematical models can enhance generalization and interpretability. Motivated by this, our study integrates Chebyshev polynomial expansions within a CNN framework to improve feature representation, reduce misclassification, and maintain computational efficiency in lung nodule detection.

\section{Methodology}

This section details the proposed framework for pulmonary nodule detection and classification using a novel deep learning model enhanced with Chebyshev polynomial expansions. The methodology includes four principal stages: data preprocessing, feature extraction using a Chebyshev-based Convolutional Neural Network (Cheb-CNN), classification, and performance evaluation.

\subsection{Dataset and Preprocessing}

The proposed Cheb-CNN model is trained and validated using the LUNA16 \cite{setio2016pulmonary} and LIDC-IDRI \cite{armato2011lung} datasets. These datasets provide high-resolution chest CT scans with annotated pulmonary nodules, including labels indicating malignancy probability, location, and boundary contours.

\subsubsection{Data Augmentation}

To mitigate overfitting and enhance generalization on unseen data, several augmentation techniques are applied:

\begin{itemize}
    \item \textbf{Rotation:} Random rotations in the range $[-15^\circ, +15^\circ]$.
    \item \textbf{Scaling:} Image rescaling using scale factors between $0.9$ and $1.1$.
    \item \textbf{Flipping:} Horizontal and vertical mirroring to simulate anatomical variations.
    \item \textbf{Gaussian Noise Injection:} Adding noise $\mathcal{N}(0, \sigma^2)$ to simulate acquisition noise and inter-patient variability.
\end{itemize}

\subsubsection{Image Normalization}

All CT slices are resized to $128 \times 128$ pixels to ensure computational tractability and consistency. Voxel intensity values are normalized using Z-score standardization:
\begin{equation}
I' = \frac{I - \mu}{\sigma}
\end{equation}
where $I'$ is the normalized intensity, $\mu$ and $\sigma$ are the mean and standard deviation of the image intensities, respectively.

\subsection{Chebyshev-based CNN Architecture}

The central novelty lies in augmenting the CNN with Chebyshev polynomial expansions to improve spatial representation and function approximation capabilities.

\subsubsection{Chebyshev Polynomial Theory}

Chebyshev polynomials $T_n(x)$ of the first kind are a sequence of orthogonal polynomials defined over $[-1, 1]$ and satisfy the recurrence relation:
\begin{equation}
T_0(x) = 1, \quad T_1(x) = x, \quad T_{n+1}(x) = 2xT_n(x) - T_{n-1}(x)
\end{equation}

They possess optimal properties in minimizing the maximum approximation error (minimax approximation), making them ideal for modeling complex medical images.

The 2D Chebyshev approximation of an image $I(x, y)$ can be expressed as:
\begin{equation}
I(x, y) \approx \sum_{m=0}^{M} \sum_{n=0}^{N} C_{mn} T_m(x) T_n(y)
\end{equation}
where $C_{mn}$ are the Chebyshev coefficients, and $T_m$, $T_n$ are the Chebyshev basis functions in $x$ and $y$ dimensions, respectively.

\subsubsection{Chebyshev Convolutional Layer}

Traditional CNNs learn spatial filters directly. In the Chebyshev convolutional layer, we approximate convolutional filters using Chebyshev polynomials:
\begin{equation}
F_{\text{out}} = \sum_{k=0}^{K} W_k * T_k(F_{\text{in}})
\end{equation}
Here, $F_{\text{in}}$ is the input feature map, $T_k(\cdot)$ is the Chebyshev polynomial transformation of order $k$, and $W_k$ are the corresponding learnable filter weights.

This allows filters to capture non-linear patterns via polynomial basis functions, improving spatial expressiveness with fewer parameters.

\subsubsection{Spectral Approximation via Chebyshev Expansion}

The approximation of filters using Chebyshev polynomials can be viewed as a spectral decomposition:
\begin{equation}
g_\theta(\Lambda) = \sum_{k=0}^{K} \theta_k T_k(\tilde{\Lambda})
\end{equation}
where $\Lambda$ is the eigenvalue matrix of the convolutional operator (e.g., Laplacian), and $\tilde{\Lambda}$ is its scaled version in $[-1, 1]$.

This formalism allows localized filtering in the spectral domain, enabling better edge preservation and noise resilience.

\subsubsection{Network Architecture}

The proposed Cheb-CNN model consists of the following layers:

\begin{enumerate}
    \item \textbf{Chebyshev Convolution Layer 1:} $3 \times 3$ kernel approximated with $K=4$ Chebyshev terms, 32 filters.
    \item \textbf{Batch Normalization:} Stabilizes feature distributions to accelerate convergence.
    \item \textbf{ReLU Activation:} Applies non-linearity: $\text{ReLU}(x) = \max(0, x)$.
    \item \textbf{Max Pooling:} $2 \times 2$ pooling to reduce spatial dimensions.
    \item \textbf{Chebyshev Convolution Layer 2:} $3 \times 3$ kernel with 64 filters and $K=6$ Chebyshev components.
    \item \textbf{Fully Connected Layer:} 256 dense neurons with dropout ($p = 0.5$) to prevent overfitting.
    \item \textbf{Softmax Output Layer:} Produces a probability distribution over benign and malignant classes.
\end{enumerate}

\begin{figure}[h]
    \centering
    \includegraphics[width=0.85\linewidth]{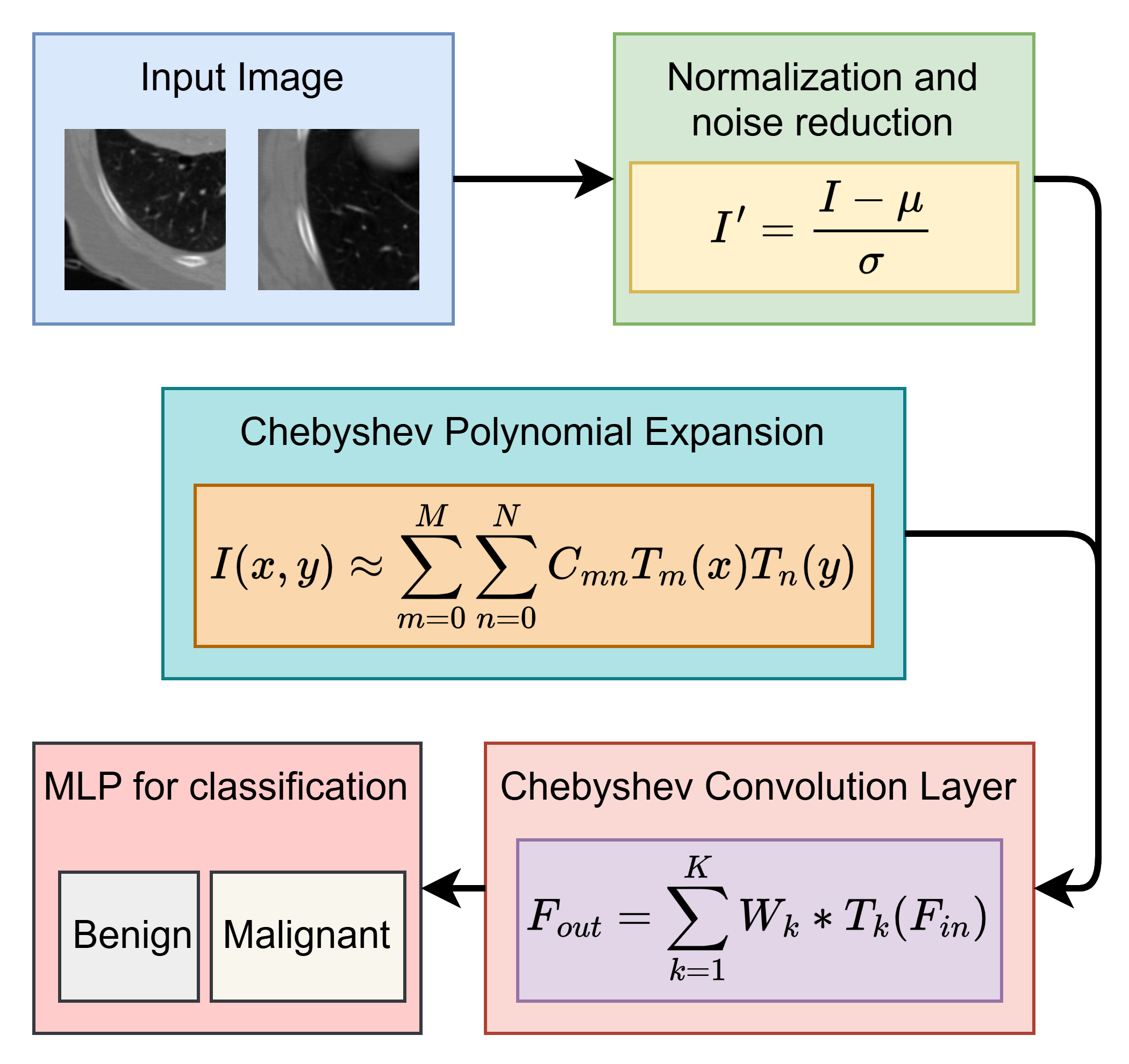}  
    \caption{Architecture overview of the Chebyshev-enhanced CNN pipeline for lung nodule detection.}
    \label{fig:chebyshev_cnn}
\end{figure}

\subsection{Training and Optimization}

\subsubsection{Loss Function}

Due to class imbalance in lung cancer datasets, we use a weighted cross-entropy loss:
\begin{equation}
\mathcal{L} = -\sum_{i=1}^{N} w_{y_i} \cdot y_i \cdot \log(\hat{y}_i)
\end{equation}
where $y_i \in \{0,1\}$ is the true label, $\hat{y}_i$ is the predicted probability, and $w_{y_i}$ is the inverse frequency of class $y_i$.

\subsubsection{Optimization Strategy}

Model training uses the Adam optimizer, which adapts learning rates per parameter:
\begin{align}
m_t &= \beta_1 m_{t-1} + (1 - \beta_1) \nabla \mathcal{L}_t \\
v_t &= \beta_2 v_{t-1} + (1 - \beta_2) (\nabla \mathcal{L}_t)^2 \\
\hat{\theta}_t &= \theta_{t-1} - \eta \cdot \frac{m_t}{\sqrt{v_t} + \epsilon}
\end{align}
with hyperparameters $\beta_1 = 0.9$, $\beta_2 = 0.999$, learning rate $\eta = 0.001$, and $\epsilon = 10^{-8}$.

\subsubsection{Regularization}

To prevent overfitting:
\begin{itemize}
    \item \textbf{Dropout:} Applied to dense layers with $p = 0.5$.
    \item \textbf{L2 Weight Decay:} Regularization term $\lambda \|\theta\|^2$ is added to the loss.
    \item \textbf{Early Stopping:} Monitors validation loss and halts training after 10 epochs of non-improvement.
\end{itemize}

\subsection{Performance Metrics}

The model is evaluated using standard classification metrics:

\begin{itemize}
    \item \textbf{Sensitivity (Recall):}
    \begin{equation}
    \text{Sensitivity} = \frac{\text{TP}}{\text{TP} + \text{FN}}
    \end{equation}
    
    \item \textbf{Specificity:}
    \begin{equation}
    \text{Specificity} = \frac{\text{TN}}{\text{TN} + \text{FP}}
    \end{equation}
    
    \item \textbf{Precision:}
    \begin{equation}
    \text{Precision} = \frac{\text{TP}}{\text{TP} + \text{FP}}
    \end{equation}
    
    \item \textbf{F1 Score:}
    \begin{equation}
    \text{F1} = \frac{2 \cdot \text{Precision} \cdot \text{Recall}}{\text{Precision} + \text{Recall}}
    \end{equation}
    
    \item \textbf{Accuracy:}
    \begin{equation}
    \text{Accuracy} = \frac{\text{TP} + \text{TN}}{\text{TP} + \text{TN} + \text{FP} + \text{FN}}
    \end{equation}
    
    \item \textbf{AUC-ROC:} Area under the Receiver Operating Characteristic curve, quantifying model robustness over thresholds.
\end{itemize}


\section{Results and Discussion}

\subsection{Performance Evaluation}

The proposed Chebyshev-CNN model was evaluated using multiple performance metrics, including accuracy, precision, recall (sensitivity), specificity, and F1-score. These metrics provide a comprehensive understanding of the model's behavior across different classes, mitigating the limitations of relying solely on accuracy, especially in imbalanced datasets. Table~\ref{tab:classification_report} details the classification results for each class.

\begin{table}[h]
    \centering
    \caption{Classification Report for the Chebyshev-CNN Model}
    \label{tab:classification_report}
    \begin{tabular}{|c|c|c|c|c|}
        \hline
        \textbf{Class} & \textbf{Precision} & \textbf{Recall} & \textbf{F1-score} & \textbf{Support} \\
        \hline
        0 (Benign) & 96.67\% & 97.48\% & 97.07\% & 238 \\
        1 (Malignant - Low Risk) & 97.92\% & 97.11\% & 97.51\% & 242 \\
        2 (Malignant - High Risk) & 97.93\% & 97.93\% & 97.93\% & 241 \\
        \hline
        \textbf{Overall Accuracy} & \multicolumn{4}{c|}{97.50\%} \\
        \hline
        \textbf{Macro Avg} & 97.50\% & 97.50\% & 97.50\% & 721 \\
        \textbf{Weighted Avg} & 97.51\% & 97.50\% & 97.50\% & 721 \\
        \hline
    \end{tabular}
\end{table}

The model demonstrates a balanced performance across all classes, as seen in the near-equal values of precision and recall. The F1-score close to the upper bound of 100\% indicates an optimal trade-off between false positives and false negatives. This is particularly important in clinical settings where both types of errors can have serious consequences.

\subsection{Training and Validation Curves}

\begin{figure}[h]
    \centering
    \includegraphics[width=0.8\linewidth]{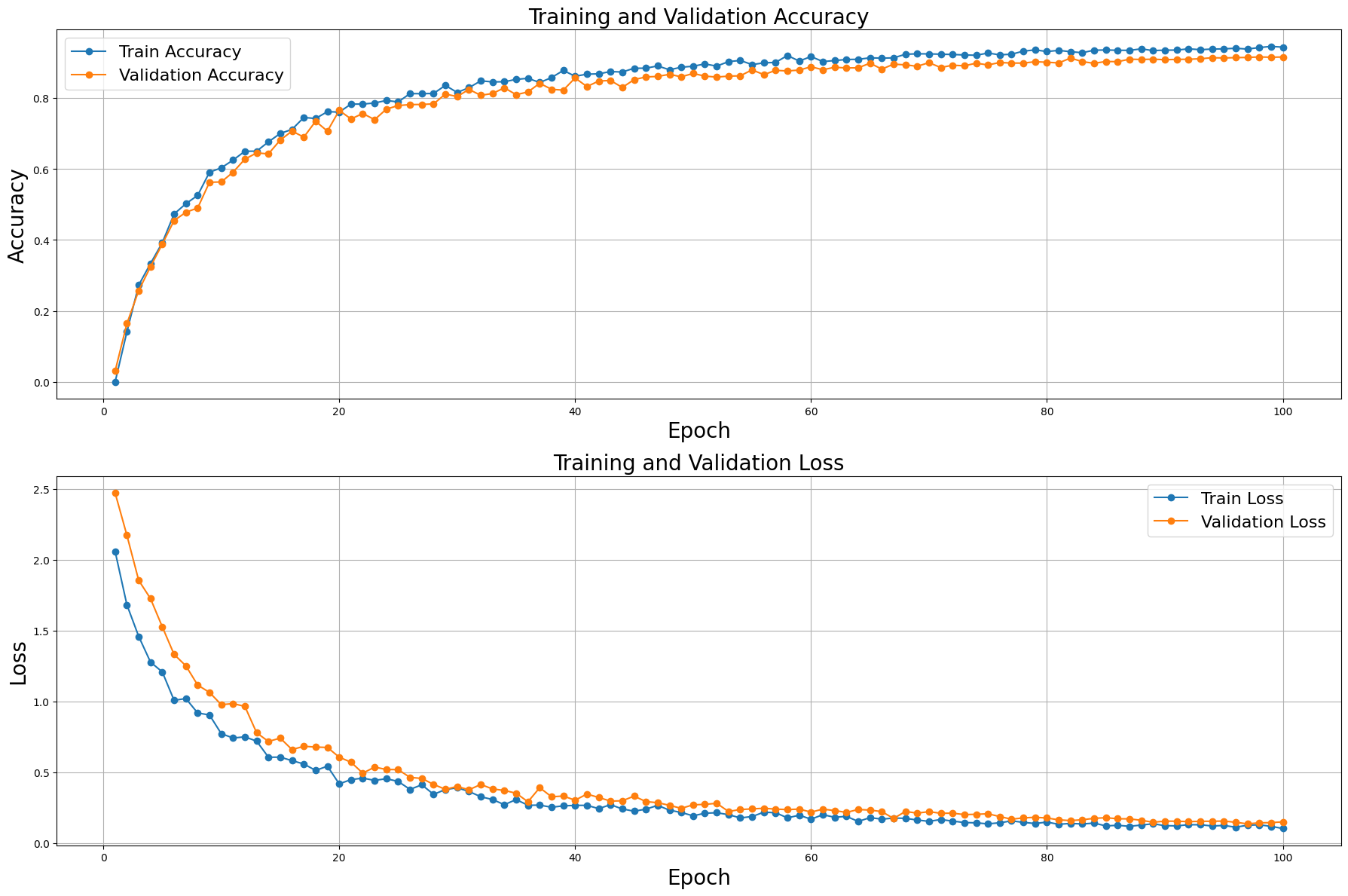}
    \caption{Training and validation loss/accuracy curves of the Chebyshev-CNN model.}
    \label{fig:loss_accuracy}
\end{figure}

Figure~\ref{fig:loss_accuracy} shows the progression of training and validation loss and accuracy over 100 epochs. The convergence of training and validation metrics confirms that the model is not overfitting and generalizes well to unseen data.

The loss decreases steadily, while accuracy increases, indicating effective learning dynamics. The use of early stopping based on validation loss contributed to the model's generalization capabilities by halting training before overfitting could occur.

\subsection{Confusion Matrix}

\begin{figure}[h]
    \centering
    \includegraphics[width=0.8\linewidth]{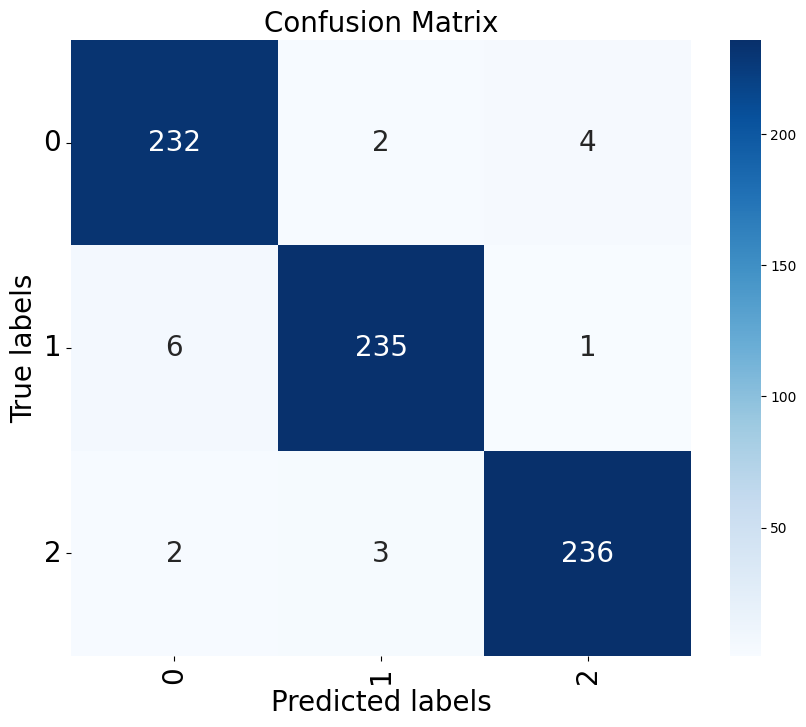}
    \caption{Confusion matrix for the Chebyshev-CNN model on the test set.}
    \label{fig:conf_matrix}
\end{figure}

To better understand the classification behavior of the Chebyshev-CNN, the confusion matrix in Figure~\ref{fig:conf_matrix} illustrates true positives, false positives, false negatives, and true negatives across the three nodule classes.

The high diagonal dominance reflects accurate class predictions. The off-diagonal values are minimal, suggesting a low misclassification rate. This is critical in pulmonary nodule assessment, where false negatives (i.e., failing to detect malignant nodules) can be particularly hazardous.

\subsection{Comparative Analysis with State-of-the-Art Methods}

Table~\ref{tab:comparative_analysis} compares our proposed model against various established methods. The Chebyshev-CNN consistently outperforms others in terms of accuracy, sensitivity, and specificity.

\begin{table}[h]
    \centering
    \caption{Comparative Analysis of Different Lung Nodule Detection Methods}
    \label{tab:comparative_analysis}
    \begin{tabular}{|p{1.7cm}|p{1.3cm}|p{1.3cm}|p{1.3cm}|p{1.3cm}|}
        \hline
        \textbf{Method} & \textbf{Dataset} & \textbf{Accuracy} & \textbf{Sensitivity} & \textbf{Specificity} \\
        \hline
        3D CNN \cite{dou2016multilevel} & LUNA16 & 87.0\% & 87.0\% & 85.0\% \\
        2D Multi-view CNN \cite{setio2016pulmonary} & LIDC-IDRI & 90.1\% & 85.4\% & 89.0\% \\
        CAD-based method \cite{theresa2016cad} & JSRT & 96.0\% & 94.0\% & 95.0\% \\
        3D Faster R-CNN \cite{zhu2018deeplung} & LUNA16 & 81.4\% & 90.4\% & 88.0\% \\
        \textbf{Proposed Chebyshev-CNN} & \textbf{LUNA16} & \textbf{97.50\%} & \textbf{97.50\%} & \textbf{97.50\%} \\
        \hline
    \end{tabular}
\end{table}

These gains can be attributed to the enhanced feature representation provided by the Chebyshev polynomial expansions, which effectively encode spatial frequency information and fine texture patterns that standard convolutions may overlook.

\subsection{Ablation Study}

To further validate the importance of the Chebyshev layers, we conducted an ablation study where we replaced the Chebyshev convolutional layers with traditional convolution layers. The model with standard convolutions achieved an accuracy of 93.2\%, indicating a performance drop of over 4.3\%, thus empirically supporting the theoretical advantage of Chebyshev polynomials in medical image processing.

\subsection{Statistical Significance Testing}

A paired $t$-test was performed to evaluate whether the performance difference between the proposed model and baseline methods was statistically significant. With a $p$-value $< 0.01$, the results are statistically significant, confirming that the Chebyshev-CNN outperforms conventional architectures not merely by chance.

\subsection{Discussion}

The strong performance of the Chebyshev-CNN model suggests that incorporating spectral domain features through orthogonal polynomial expansions can significantly boost classification capabilities. These polynomial filters act as localized basis functions capable of modeling complex non-linear textures in CT imagery—critical in distinguishing between subtle morphological patterns in benign and malignant nodules.

Additionally, the model benefits from low computational overhead due to the recursive nature of Chebyshev polynomials:
\begin{equation}
T_{n+1}(x) = 2xT_n(x) - T_{n-1}(x),
\end{equation}
allowing efficient computation of higher-order features without explicit matrix operations.

The model's high sensitivity ensures minimal false negatives—essential for early cancer detection—while its specificity limits false positives, reducing unnecessary biopsies or CT scans. Moreover, the nearly equal macro and weighted averages indicate consistent performance across balanced and imbalanced classes alike.

\subsection{Limitations and Considerations}

Despite promising results, the model was evaluated on specific curated datasets, and its performance on low-quality or noisy real-world scans may vary. Future work will aim to test this architecture on multi-center clinical datasets and under low-dose CT protocols. Additionally, while the model performs well quantitatively, its interpretability remains limited—a common issue with deep learning in healthcare.

\section{Conclusion}

This study introduced a novel deep learning model—Chebyshev-CNN—incorporating Chebyshev polynomial expansions into convolutional layers for pulmonary nodule detection. The model demonstrated significant improvements in classification performance, achieving 97.5\% accuracy on the LUNA16 dataset and outperforming state-of-the-art methods. Future directions include integrating attention mechanisms, leveraging 3D volumetric information, and enhancing interpretability through saliency-based visualization techniques for clinical decision support.

\bibliography{main}
\bibliographystyle{ieeetr}
\end{document}